\documentclass[letterpaper]{article} 
\usepackage{aaai24}  
\usepackage{times}  
\usepackage{helvet}  
\usepackage{courier}  
\usepackage[hyphens]{url}  
\usepackage{graphicx} 
\usepackage{natbib}  
\usepackage{caption} 
\usepackage{makecell}
\usepackage{amsmath,amssymb,amsfonts}
\usepackage{amsthm}
\usepackage{multirow}
\usepackage{caption}
\usepackage{subfigure}

\usepackage[framemethod=tikz]{mdframed}
\definecolor{mygray}{RGB}{243,243,244}
\newmdenv[
  innertopmargin=0pt,
  backgroundcolor=mygray,
  linecolor=none,
  innerleftmargin=0pt,
  innerrightmargin=0pt,
  leftmargin=0pt
  ]{mymath}

\usepackage[ruled, linesnumbered,vlined]{algorithm2e}

\title{Motif-aware Riemannian Graph Neural Network with \\Generative-Contrastive Learning}
\author{
    Li Sun\textsuperscript{\rm 1}\thanks{Corresponding Authors: Li Sun and Hao Peng},
    Zhenhao Huang\textsuperscript{\rm 1},
    Zixi Wang\textsuperscript{\rm 1},
    Feiyang Wang\textsuperscript{\rm 2},
    Hao Peng\textsuperscript{\rm 3$\ast$},
    Philip Yu\textsuperscript{\rm 4}
}
\affiliations{
    \textsuperscript{\rm 1}North China Electric Power University, Beijing 102206, China\\
    \textsuperscript{\rm 2}Beijing University of Posts and Telecommunications, Beijing 100876, China\\
    \textsuperscript{\rm 3}Beihang University, Beijing 100191, China\\
    \textsuperscript{\rm 4}Department of Computer Science, University of Illinois at Chicago, IL 60607, USA\\
    E-mails: ccesunli@ncepu.edu.cn; penghao@buaa.edu.cn; psyu@uic.edu
}

\begin{document}

\maketitle

\begin{abstract}
Graphs are typical non-Euclidean data of complex structures. 
In recent years, Riemannian graph representation learning has 
emerged as an exciting alternative to Euclidean ones. 
However, Riemannian methods are still in an early stage: 
most of them 
present a single curvature (radius) regardless of structural complexity, 
suffer from numerical instability due to the exponential/logarithmic map,
and lack the ability to capture motif regularity. 
In light of the issues above, we propose the problem of \emph{Motif-aware Riemannian Graph Representation Learning}, 
seeking a numerically stable encoder to capture motif regularity in a diverse-curvature manifold without labels.
To this end, we present a novel Motif-aware Riemannian model with Generative-Contrastive learning (MotifRGC),
which conducts a minmax game in  Riemannian  manifold in a self-supervised manner.
First, we propose a new type of Riemannian GCN (D-GCN), in which we construct a diverse-curvature manifold by a product layer with the diversified factor,
and replace the exponential/logarithmic map by a stable kernel layer.
Second, 
we introduce a motif-aware Riemannian generative-contrastive learning 
to capture motif regularity in the constructed manifold and learn motif-aware node representation without external labels.
Empirical results show the superiority of MofitRGC.
\end{abstract}


\section{Introduction}

Graphs are the natural descriptions of real systems, ranging from social networks and recommender systems to chemistry and bioinformatics.
Graph representation learning shows fundamental importance in a variety of learning tasks, such as node classification and link prediction \cite{hamilton2018inductive,velickovic2018graph}.
Euclidean space has been the workhorse of graph representation learning for decades \cite{kdd14DeepWalk,kipf2017semisupervised}.
It is not until recently that \textbf{Riemannian spaces} have emerged as an exciting alternative
\cite{iclr21hnn++,www22curvGAN,iclr22understanding,icml18revisitHyperbolic},
 since they generally better match the geometry of the graph than the  Euclidean counterpart \cite{petersen2016riemannian,DBLP:journals/corr/abs-1006-5169}.
While achieving encouraging performance, Riemannian graph representation learning is still in its early stages.
There are several important issues largely remaining open.

The first issue is on the \emph{curvature diversity}. 
Most of  prior Riemannian methods \cite{zhang2021lorentzian,dai2021hyperbolictohyperbolic,liu2019hyperbolic} study graphs in the manifold of a single curvature (radius), which is  only suitable for a special type of graphs. 
For example, a negative curvature manifold (hyperbolic space) is well aligned with hierarchical and tree-like graphs \cite{nips17NickelK,chen2022fullya}.
A positive curvature manifold (hyperspherical space) is suitable for cyclical graphs \cite{bachmann2020constant}.
Note that, the product of single-curvature manifolds \cite{nips19product} still presents a single curvature as a whole.
The recent quotient manifold \cite{xiong2022pseudoriemannian} also relies on a single curvature radius.
\emph{There is a lack of curvature diversity to model the complex structures of real graphs.}

\begin{table}[t]
\caption{Comparison of proposed MotifRGC and previous Riemannian methods. ($\kappa$ denotes the curvature.)}
\vspace{-0.1in}
\resizebox{1\linewidth}{!}{ 
\begin{tabular}{r  | c c  c c }
         &{\makecell[c]{Curvature\\ Diversity}} & {\makecell[c]{Motif\\ Regularity}} & {\makecell[c]{Numerical\\ Stability}}    \\
\hline
HGCN (NeurIPS: \citeyear{chami2019hyperbolica}) & one $\kappa<0$ & &  \\
$\kappa$-GCN (ICML: \citeyear{bachmann2020constant})& one $\kappa\in \mathbb R$ & &  \\
LGCN (WWW: \citeyear{zhang2021lorentzian}) & one $\kappa<0$ & &  \\
H-to-H (CVPR: \citeyear{dai2021hyperbolictohyperbolic}) & one $\kappa<0$& &\checkmark  \\
\emph{fully}H (ACL: \citeyear{chen2022fullya}) & one $\kappa<0$  & &\checkmark \\
$Q$-GCN (NeurIPS: \citeyear{xiong2022pseudoriemannian}) & one radius $\frac{1}{|\kappa|}$  & &  \\
\hline
\textbf{MotifRGC (Ours)}& \emph{diverse}  &\checkmark &\checkmark  \\
\hline
\end{tabular} 
}
\label{compare}
\vspace{-0.2in}
\end{table}

The second issue is on the \emph{numerical stability}. 
Riemannian methods \cite{chami2019hyperbolica,liu2019hyperbolic,xiong2022pseudoriemannian} typically model graphs with the couple of exponential and logarithmic maps.
Unfortunately, exponential/logarithmic map is not strictly numerical-stable, and the issue of ``Not a Number'' (NaN) may occur in practice without numerical processing or careful hyperparameter tuning \cite{chen2022fullya,Yu2022RandomLF}.
Recently, \citet{dai2021hyperbolictohyperbolic} introduce the Lorentz-type operators to replace exponential/logarithmic map, but pose difficulty in optimization 
as side effect.
Thus, \emph{the issue of numerical stability largely remains open}.

The third issue is on the \emph{motif regularity}. 
The motifs (small substructures such as triangles  and cliques) are the fundamental building blocks of a graph, 
and they play an essential role in modeling and understanding social or biochemical networks \cite{HostPool,DBLP:conf/icml/YuG22}. 
Thus, encoding motif regularity is important in graph representation learning, and benefits downstream tasks, e.g., node classification and link prediction.
In the literature, motif regularity has been extensively studied in Euclidean space \cite{DBLP:conf/kdd/LiuS23,DBLP:conf/aaai/Subramonian21}.
Riemannian manifold tends to be more suitable to model motifs than the Euclidean counterpart, as will be shown in the experiment.
Surprisingly, \emph{it has been rarely explored in generic Riemannian manifolds, to our best knowledge.}


The aforementioned issues motivate us to rethink Riemannian graph representation learning, and propose a new problem of 
 \emph{Motif-aware Riemannian Graph Representation Learning}, which aims at finding a numerically stable graph encoder to model motif regularity in a novel diverse-curvature manifold.
Besides, \emph{self-supervised learning} without external labels is more practical, as labeling graphs is expensive or even impossible.
In Riemannian manifolds, \citet{www22DualSpace,aaai22SelfMix} leverage contrastive learning recently, while generative learning is still under investigated.

\textbf{Our approach.} To this end, we propose 
a novel Motif-aware Riemannian model with Generative-Contrastive Learning (\textbf{MotifRGC}), which conducts a minmax game in Riemannian manifold in a self-supervised manner. 
First, we propose a new type of GCN, namely Diverse-curvature GCN (D-GCN).
In our design, we construct a diverse-curvature manifold by a product layer, where a \emph{diversified factor} is introduced to ensure the curvature diversity (issue one).
We replace the exponential/logarithmic map suffering from numerical stability (issue two) by a gyrovector \emph{kernel layer}, where a numerical stable map based on Bochner’s Theorem is formulated.
Second, we introduce the motif-aware Riemannian generative-contrastive learning, exploring the duality of the constructed manifold.
In the product manifold, we design a \emph{Riemannian motif generator} to generate fake motifs. The curvatures are learned to capture motif regularity (issue three), when the generated motifs become indistinguishable to the discriminator.
Among the factor manifolds, node representations are learned by contrasting different geometric views of the factors, in which we introduce a \emph{motif-aware hardness} to highlight the hard samples. 
We compare our model with prior methods in Table 1.

Notable contributions are summarized as follows:
\begin{itemize}
    \item  \emph{Problem.} We make the first attempt to study motif-aware Riemannian graph representation learning, encoding motif regularity in a diverse-curvature manifold.
   \item  \emph{Methodology.}  In MotifRGC, we propose a novel D-GCN coupled with several theoretical guarantees. 
   Furthermore, we introduce the motif-aware Riemannian generative-contrastive learning to generate motif-aware node representations in a self-supervised manner.
   \item  \emph{Experiment.} Empirical results show MotifRGC outperforms previous Riemannian models. Codes are given in \url{https://github.com/RiemannGraph/MotifRGC}. 
\end{itemize}

\section{Preliminaries}
We introduce necessary mathematics, specify the limitations of prior methods and propose the studied problem.

\subsection{Riemannian Geometry}

\noindent\textbf{Manifold $\ $}
A Riemannian manifold $\mathbb M$ is a smooth manifold coupled with a Riemannian metric.
For each point $\boldsymbol x$, the Riemannian metric $g_{\boldsymbol x}$ is defined on its tangent space $\mathcal T_{\boldsymbol x}\mathbb M$.
The logarithmic map $\operatorname{Log}_{\boldsymbol x}: \mathbb M \to \mathcal T_{\boldsymbol x}\mathbb M$ transforms the vector in the manifold to the tangent space, while the exponential map $\operatorname{Exp}_{\boldsymbol x}$ does the inverse transform.
Euclidean space is a special case of Riemannian manifold.

\noindent\textbf{Curvature $\ $}  Curvature measures the extent how a surface derives from being flat, and determines the shape of manifold. 
Each point $\boldsymbol x$ in the manifold is associated with a curvature $\kappa_{\boldsymbol x}$ and a corresponding curvature radius $\frac{1}{|\kappa_{\boldsymbol x}|}$.
A manifold is said to be a \emph{single-curvature manifold} when the curvature (radius) of each point is equal.
Specifically, the manifold is hyperbolic $\mathbb H$ if the single curvature is negative, while the manifold is hyperspherical $\mathbb S$ for a positive curvature.
On the contrary, \emph{diverse-curvature manifold} refers to a manifold where the curvatures of its points are not the same.
Most previous works study graph in a single-curvature manifold \cite{DBLP:conf/kdd/00010PK23,chami2019hyperbolica,xiong2022pseudoriemannian}.
\textbf{In fact, it calls for a diverse-curvature manifold, better matching the complex structures of real graphs.}

\subsection{Euclidean \& Riemannian GCNs}

Graph Convolutional Networks (GCNs) are the dominant method for graph representation learning. 
GCNs typically conduct message passing over graphs \cite{velickovic2018graph,hamilton2018inductive}.
Concretely, in the convolution layer, each node representation $\boldsymbol h_i$ aggregates the information of neighboring nodes $\mathcal N_i$ and  combines the aggregated information to itself. 
In Euclidean space, it takes the form of
\begin{equation}
\bar{\boldsymbol h}_i=\operatorname{Agg}_{j \in \mathcal N_i}(\boldsymbol h_j), \quad  \boldsymbol h_i=\operatorname{Comb}(\bar{\boldsymbol h}_i, \boldsymbol h_i),  
\end{equation}
where $\operatorname{Agg}$ and $\operatorname{Comb}$ denote aggregation and combination, respectively.
Similarly, a Riemannian GCN layer is generally formulated as follows, 
\begin{equation}
\resizebox{0.909\hsize}{!}{$
\bar{\boldsymbol h}_i=\operatorname{Agg}_{j \in \mathcal N_i}(\operatorname{Log}_{\mathbf 0}(\boldsymbol h_j)),  \boldsymbol h_i=\operatorname{Exp}_{\mathbf 0}(\operatorname{Comb}(\bar{\boldsymbol h}_i, \boldsymbol h_i)),
$}
\label{exp-log-layer}
\end{equation}
where $\mathbf 0$ denotes the reference point of exponential and logarithmic maps.
However, as shown in \citet{chen2022fullya,Yu2022RandomLF}, \textbf{the exponential/logarithmic map poses the issue of numerical stability.}

\begin{figure*}
\centering
    \includegraphics[width=1\linewidth]{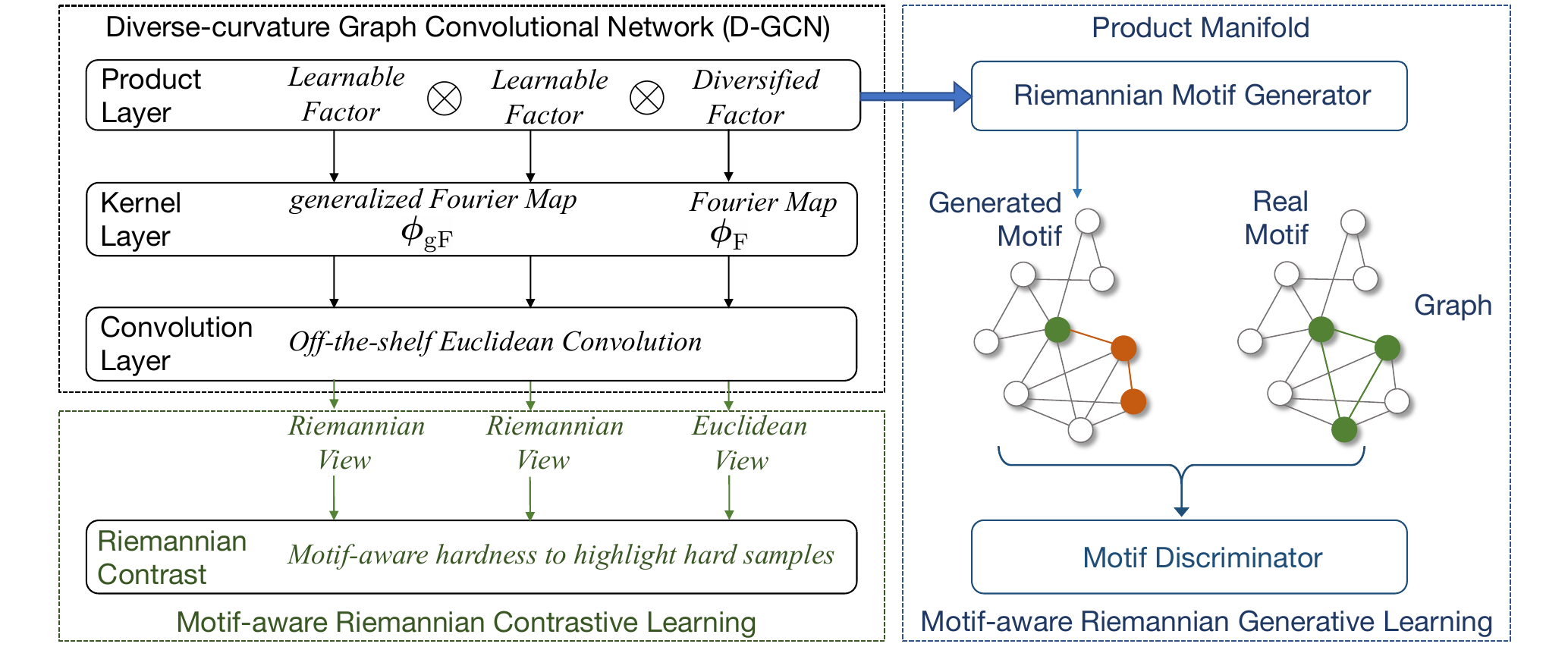}
    \vspace{-0.25in}
     \caption{Overall architecture of \textbf{MotifRGC}. The proposed D-GCN consists of product layer, kernel layer and convolution layer. We learn the curvatures of the product manifold in the motif-aware Riemannian generative learning (\textcolor{blue}{\textbf{blue box}}),  and learn the node representations in the motif-aware Riemannian contrastive learning (\textcolor{olive}{\textbf{olive box}}).}
    \label{illu}
        \vspace{-0.1in}
\end{figure*}

\subsection{Problem Formulation}

\subsubsection{Notations.} 
Lowercase $\boldsymbol x$ and italic uppercase $\mathcal X$ denote vector and set, respective.
$\|\cdot\|$ denotes the usual $L2$ norm.

A graph is described as a tuple of $G(\mathcal V, \mathcal E, \mathcal M, \mathbf X)$,  where $\mathcal V=\{v_1, \cdots, v_N\}$ is the node set and $\mathcal E=\{(v_i, v_j)\} \subset \mathcal V \times \mathcal V$ is the edge set.
$\mathbf X \in \mathbb R^{N \times F}$ is the feature matrix, whose $i^{\operatorname{th}}$ row is the Euclidean feature of node $v_i$.
Motifs $\mathcal M=\{M^3, \cdots, M^K\}$ are small subgraphs frequently occurred in the graph, such as triangles and cliques. 
A motif $M^K$ is the set of $K-$node connected subgraphs, where motif node set is $\mathcal V^K\subset \mathcal V$,   $|\mathcal V^K|=K$, and motif edge set is $\mathcal E^K\subset \mathcal E$.
We observe that \textbf{the motifs have not yet been explored in Riemannian manifold of diverse curvatures.}
Also, we notice that labeling graph is often infeasible in practice. 
Thus, we propose a new problem as follows.
\newtheorem*{def1}{Problem Definition}
\begin{def1}
 \emph{(Motif-aware Riemannian Graph Representation Learning.)} Given a graph $G(\mathcal V, \mathcal E, \mathcal M, \mathbf X)$ without any external labels, the proposed problem aims to learn a \underline{numerically stable} encoding function $\Phi: \mathcal V \to \mathbb M$, so that \underline{motif regularity} is captured in a \underline{diverse-curvature manifold} $\mathbb M$, matching the complex structures of $G$.
\end{def1}

\noindent In short, we rethink Riemannian graph representation learning, and our work is distinguished from the prior works on curvature diversity, motif regularity and numerical stability.

\section{Our Approach}

To address this problem, we propose a novel self-supervised Motif-aware Riemannian model with Generative and Contrastive learning (\textbf{MotifRGC}).
In a nutshell,
we first propose a \emph{new type of Riemannian GCN} to address the issues of curvature diversity and numerical stability, namely D-GCN.
Then, we propose \emph{Motif-aware Riemannian Generative-Contrastive Learning} to capture motif regularity via a minmax game in  Riemannian manifold.
The overall architecture is illustrated in Figure 1, and we elaborate on our D-GCN and learning approach in Sec 3.1 and Sec 3.2, respectively.

\subsection{A New Formulation of D-GCN}

Different from typical Riemannian GCNs in Eq. (\ref{exp-log-layer}), we propose a novel Diverse-curvature GCN (termed as D-GCN), where
we construct a diverse-curvature manifold  via a product layer with a diversified factor,
and replace the exponential/logarithmic map via a kernel layer. 
Another novelty lies in that the curvature learning of the manifold is conducted  in a motif generation process, detailed in Sec. 3.2.

\subsubsection{Product Layer}

The product layer conducts Cartesian product of multiple \emph{learnable} single-curvature factors and one diversified factor.
A \textbf{single-curvature factor} is a Riemannian manifold of $\kappa-$stereographical model $\mathbb G_{\kappa}^d$ \cite{petersen2016riemannian},
where $\kappa$ and $d$ denote curvature and dimension, respectively.
Specifically, $\mathbb G_{\kappa}^d$ is defined on a smooth gyrovector ball  $\left\{\boldsymbol{x} \in \mathbb{R}^{d} \mid-\kappa \|\boldsymbol{x}\|^{2}<1\right\}$
with distance metric of $d(\boldsymbol x, \boldsymbol y)=\frac{2}{\sqrt{|\kappa|}}\tan^{-1}_\kappa\left(\sqrt{|\kappa|} \|-\boldsymbol x \oplus_\kappa \boldsymbol y\|\right)$,
where gyrovector addition $\oplus_\kappa$ and curvature trigonometry $\tan^{-1}_\kappa$ are detailed in the Appendix.
The merit of $\kappa-$stereo- graphical model is that it unifies hyperbolic and hyperspherical spaces in gyrovector formalism.
The \textbf{diversified factor} is an upper hypersphere $\mathbb S^{d_0}_{\operatorname{up}}$, which is responsible to diversify curvature of the product manifold. 
$\mathbb S^{d_0}_{\operatorname{up}}$ is expressed in the polar coordinates of $(r, \boldsymbol \theta)$, where $r$ and $\boldsymbol \theta$ denote the magnitude coordinate and angular vector, respectively.
Only $r-$coordinate contributes to curvature and distance, detailed in Appendix, and we set the norm of Euclidean feature as $r$.

The resulting product manifold is derived as follows
\begin{equation}
\mathbb M=\mathbb G^{d_1}_{\kappa_1} \otimes \mathbb G^{d_2}_{\kappa_2} \otimes\cdots \otimes \mathbb G^{d_M}_{\kappa_M} \otimes \mathbb S^{d_0}_{\operatorname{up}},
\label{product}
\end{equation}
where $\otimes$ denotes the Cartesian product. 
A point in the product manifold takes the form $\boldsymbol x=[\boldsymbol x^1|| \cdots || \boldsymbol x^M || \boldsymbol x^0] \in \mathbb M$,
where $||$ is vector concatenation.  $\boldsymbol x^m \in \mathbb G^{d_m}_{\kappa_m}$ is the $m^{\operatorname{th}}$ component of $\boldsymbol x$, and $\boldsymbol x^0 \in \mathbb S_{\operatorname{up}}^{d_0}$.
The distance metric is 
\begin{equation}
\resizebox{0.9\hsize}{!}{$
d^2_{\mathbb M}(\boldsymbol x, \boldsymbol y)= \sum\nolimits_{m=1}^M d^2_{\kappa_m}(\boldsymbol x^m, \boldsymbol y^m) + \left(\|\boldsymbol x^0\|-\| \boldsymbol y^0\|\right)^2.
$}
\label{dist}
\end{equation}
With Eqs. (\ref{product})$-$(\ref{dist}), we have Proposition 1 hold.
\begin{mymath}
\newtheorem*{pro1}{Proposition 1 (Curvature Diversity)} 
\begin{pro1}
The manifold $\mathbb M$ constructed by the product layer (Eq. \ref{product}) is a diverse-curvature manifold in which each point has its own curvature $\kappa_{\boldsymbol x}$ determined by its location in the manifold.
\end{pro1}
\begin{proof}
Please refer to the Appendix.
\end{proof}
\end{mymath}

\subsubsection{Kernel Layer}



Recall that, typical Riemannian GCN layer with exponential/logarithmic map suffers from the issue of numerical stability.
Instead, we put forward a fresh perspective of kernel method:
We transform Riemannian features to Euclidean ones via \emph{a gyrovector kernel}, 
and then utilize well-established Euclidean GCN layers  to update node representations, as shown in Fig. 1.

The transformation between two types of spaces is challenging due to the difference in geometry and constraint on gyrovector ball. 
In kernel layer, we formulate a \textbf{generalized Fourier Map} $\boldsymbol \phi_{\operatorname{gF}}: \mathbb G_{\kappa}^{n} \to \mathbb R^{m}$ to address this challenge.
\emph{First, we start with Fourier (feature) map $\boldsymbol \phi_{\operatorname{F}}:\mathbb R^{n} \to \mathbb R^{m}$ in Euclidean space.}
The Bochner’s Theorem \cite{RahimiR07-Thm} utilizes eigenfunctions to construct the Fourier map for any invariant kernel.
A typical Euclidean eigenfunction family takes the form of
\begin{equation}
\operatorname{F}_{\boldsymbol \omega, b}(\boldsymbol x)=\sqrt{2}\cos(\langle \boldsymbol \omega, \boldsymbol x\rangle +b), \boldsymbol x \in \mathbb R^{n},
 \label{fm}
\end{equation} 
where phase vector $\boldsymbol \omega$ and bias $b$ are uniformly sampled from a $n-$dimensional unit ball and $[0, 2\pi]$, respectively.

\noindent \emph{Second, we generalize Euclidean $\operatorname{F}_{\boldsymbol \omega, b}(\cdot)$ to Riemannian manifold.}
Analogous to Eq. (\ref{fm}), we derive the eigenfunction in the gyrovector ball $\mathbb G_{\kappa}^{n}$ of Riemannian manifold,
\begin{equation}
\operatorname{gF}^\kappa_{\boldsymbol \omega, b, \lambda}(\boldsymbol x)=A_{\boldsymbol \omega, \boldsymbol x}\cos\left( \lambda \langle \boldsymbol \omega, \boldsymbol x \rangle_\kappa +b \right), \boldsymbol x \in \mathbb G_{\kappa}^{n},
\label{gyroWave}
\end{equation}
where the amplitude function $A_{\boldsymbol \omega, \boldsymbol x}$ is $\exp\left(\frac{n-1}{2}\langle \boldsymbol \omega, \boldsymbol x \rangle_\kappa\right)$.
$
\langle \boldsymbol \omega, \boldsymbol x \rangle_\kappa =\log \frac{1+\kappa \| \boldsymbol x \|^2}{\| \boldsymbol x - \boldsymbol \omega \|^2}
$
is the signed distance in the gyrovector ball.
We draw $m$ independent samples of  $\boldsymbol \omega$, $b$  and $\lambda$ uniformly from $n-$dimensional gyrovector ball, $[0, 2\pi]$ and Gaussian distribution,  respectively.
The generalized Fourier map is then given as
\begin{equation}
\resizebox{1\hsize}{!}{$
\boldsymbol \phi_{\operatorname{gF}}(\boldsymbol x)=\frac{1}{\sqrt{m}}\left[\operatorname{gF}^\kappa_{\boldsymbol \omega_1,\lambda_1,b_1}(\boldsymbol x), \cdots,\operatorname{gF}^\kappa_{\boldsymbol \omega_m,\lambda_m,b_m}(\boldsymbol x) \right] \in \mathbb R^{m}.
$}
\end{equation}

Theoretically, we show that induced kernel can be regarded as a generalization of Poisson kernel \cite{icml22RidgeletKernel} to Riemannian manifold.
\begin{mymath}
\newtheorem*{pro3}{Proposition 2 (Poisson Kernel)} 
\begin{pro3}
Let the radius of gyrovector ball be $1$ (curvature $\kappa=-1$), the kernel in Proposition 2 is equivalent to the famous Poisson kernel. 
\end{pro3}
\begin{proof}
Please refer to the Appendix.
\end{proof}
\end{mymath}

\subsubsection{Network Architecture}
In our D-GCN, the product layer assigns each node with multiple learnable Riemannian features, and then constructs the diverse-curvature manifold. 
The kernel layer conducts $\boldsymbol \phi_{\operatorname{gF}}$ to Riemannian features and $\boldsymbol \phi_{\operatorname{F}}$ to Euclidean features.
The convolution layer can be given by any off-the-shelf Euclidean one.


\subsection{Motif-aware Generative-Contrastive Learning on Riemannian Manifold}

We propose a novel Motif-aware Riemannian Generative-Contrastive learning, in which \emph{we introduce a minmax game to Riemannian manifolds.}
Thanks to the formulation in Eq. (\ref{product}), the constructed manifold presents a duality of the product manifold as a whole and the collaboration of factor manifolds.
Exploring the duality of the constructed manifold,
we conduct motif generation process in the product manifold  for the curvature learning (\emph{Generative Learning}).
Simultaneously, we  contrast among factor manifolds for the representation learning (\emph{Contrastive Learning}), where a motif-aware hardness is introduced.






In generative learning, the product manifold is trained to capture motif regularity, so that \emph{fake motifs $\mathcal S$ generated  from  the generator $G$ cannot be distinguished from real motifs $\mathcal M$ by the discriminator $D$.}
Specially, generator $G$ and discriminator $D$ play the minmax game as follows,
\begin{equation}
\resizebox{1\hsize}{!}{$
\min\limits_{\theta_G} \max\limits_{\theta_D} \ \mathbb E_{\mathcal M}[\log D(\mathcal M;\theta_D)] +\mathbb E_{\mathcal S\sim G(\mathcal S;\theta_G)}[\log (1-D(\mathcal S;\theta_D))].
$}
\label{minmax}
\end{equation}

\subsubsection{Riemannian Motif Generator}
We design a Riemannian motif generator $G$ to generate indistinguishable motifs  in the minimization of Eq. (\ref{minmax}). 
A motif $\mathcal M^k$ is a set of $k$ nodes $\mathcal S=\{v_{s_1}, \cdots, v_{s_{k}}\}$.
Let $\mathcal S_{s_1}$ denote the node set including $v_{s_1}$.
Given $v_{s_1}$, the selection of other $(k-1)$ nodes is guided by the proposed generator defined as follows,
\begin{equation}
\begin{aligned}
& G(\mathcal S_{s_1}| v_{s_1})= \\ 
&G_c(v_{s_2}| v_{s_1})G_c(v_{s_3}| v_{s_1}, v_{s_2})\cdots G_c(v_{s_k}| v_{s_1}, \cdots, v_{s_{k-1}})
\end{aligned}
\label{mg1}
\end{equation}
The generation of $v_{s_k}$ is based on previously selected nodes $\mathcal S'=\{v_{s_1}, \cdots, v_{s_{k-1}}\}$, and
the probability of $v_{s_k}$ being selected is defined by \textbf{distance metric in the product} $\mathbb M$.
Concretely, we apply a softmax function over all other nodes, 
\begin{equation}
\resizebox{1\hsize}{!}{$
G_c(v_{s_k}| v_{s_1}, \cdots, v_{s_{k-1}})=\frac{\exp(d_{\mathbb M}(v_{s_k}, \mu_{{s_1}, \cdots, s_{k-1}}))}{\sum_{v_{s_j} \notin \mathcal S'} \exp(d_{\mathbb M}(v_{s_j}, \mu_{{s_1}, \cdots, s_{k-1}}))},
$}
\label{mg2}
\end{equation}
where $d_{\mathbb M}$ given in Eq. (\ref{dist}) is a function regarding the curvatures in the product.
$\mu_{{s_1}, \cdots, s_{k-1}}$ is the geometric centroid of the nodes $\mathcal S'$ in $\mathbb M$.
Thanks to combinational formulation of $d_{\mathbb M}$, there is no need to consider the centroid as a whole, and only the centroid in each factor $\mu^m_{{s_1}, \cdots, s_{k-1}}$ is required. 
For each factor of gyrovector ball $\mathbb G^{d_m}_{\kappa_m}$, it is given by the gyro-midpoint \cite{bachmann2020constant},
\begin{equation}
 \mu^m_{{s_1}, \cdots, s_{k-1}}=\sum_{i=1}^{k-1}\frac{\lambda^{\kappa_m}_{v_{s_i}}}{\sum_{j=1}^{k-1}(\lambda^\kappa_{v_{s_j}}-1)}v^m_{s_i} \  \in \mathbb G^{d_m}_{\kappa_m}, 
\end{equation}
where $\lambda^\kappa_{v}=2/(1+\kappa\| v \|^2)$ is the conformal factor.

In the minimization of Eq. (\ref{minmax}), curvatures of the product  are the parameters $\theta_G$ to be learned.
The intuition is that \emph{the product manifold is able to generate indistinguishable motifs, when its curvatures (and corresponding distance metric) capture the motif regularity in graph.}
The sampling of motifs $\mathcal S\sim G(\mathcal S;\theta_G)$ is discrete, and thus we minimize Eq. (\ref{minmax}) via the policy gradient \cite{aaai18GraphGAN} to train $\theta_G$. 
Here, we focus on a fundamental motif of triangle.

\subsubsection{Motif Discriminator}
In the maximization of Eq. (\ref{minmax}), the discriminator $D$ aims to classify the generated motifs from the real ones sampled from the graph. 
Given a node set $\{v_{s_1}, \cdots, v_{s_{k}}\}$, we implement the discriminator as follows,
\begin{equation}
\resizebox{0.88\hsize}{!}{$
D(v_{s_1}, \cdots, v_{s_{k}}| \theta_D)=\operatorname{MLP}(\operatorname{Pooling}(v_{s_1}, \cdots, v_{s_{k}})),
$}
\end{equation}
where $\operatorname{MLP}$ is a MultiLayer Perception whose output layer is the sigmoid.
We apply mean-pooling to the node representations from the kernel layer of our Riemannian GCN, so that the representations can be tackled by a normal $\operatorname{MLP}$. Thus, $\theta_D$ is the parameter of $\operatorname{MLP}$.




\subsubsection{Motif-aware Riemannian Contrast}

We formulate a minimization objective for motif-aware Riemannian contrastive learning, 
in which we introduce a motif-aware hardness to highlight hard samples for learning node representations.


First, we generate multiple geometric views, and each factor manifold provides a geometric view of corresponding geometry.
For factor manifold $\mathbb G^{d^m}_{\kappa^m}$, its geometric view is derived as  $\boldsymbol z^m=g_{\mathbf \Theta}(\boldsymbol \phi_{\operatorname{gF}}(\boldsymbol x^m))$, 
and $\boldsymbol z^0=g_{\mathbf \Theta}(\boldsymbol \phi_{\operatorname{F}}(\boldsymbol x^0))$ for Euclidean view.
$\boldsymbol x^m$ and $\boldsymbol x^0$ denote the Riemannian features and Euclidean features, respectively. 
Each geometric view is contrasted to Euclidean view and vice versa, so that positive samples are close and negative samples are pushed away.

Second, we leverage the motif to distinguish positive/ negative samples.
Intuitively, the nodes in a motif are similar to each other, and are considered as positive samples.
Take the motif of triangle $\mathcal M^3$ for instance. 
For node $v_i$, the set of positive samples is denoted as $\mathcal V^3(i)$,
which is a collection of the endpoints in triangles including node $v_i$.

Furthermore, we formulate a \textbf{motif-aware hardness} to highlight the hard samples, i.e., the positive samples with low similarity and the negative samples with high similarity.
\begin{equation}
h(\boldsymbol z_i, \boldsymbol z_j|\mathcal M_3)=|\mathbb I(j \in \mathcal V^3(i)) -\operatorname{Normal}(s(\boldsymbol z_i, \boldsymbol z_j))|^\alpha,
 \label{hard}
\end{equation}
where the indicator function $\mathbb I(\cdot)$  returns $1$ iff $(\cdot)$ is true, $s$ is a similarity measure, and $\operatorname{Normal}$ normalizes the similarity value to $[0,1]$. The positive $\alpha$ controls the effect of hardness.
\emph{Eq. (\ref{hard}) up-weights the hard samples.} 
For example, given $\alpha=2$, a hard negative sample with similarity $s=0.9$ is reweighed by $0.81$, while an easy negative with $s=0.1$ is reweighed by $0.01$.
The reweighing of motif-aware hardness pays more attention to the hard samples than the typical equal treatment, i.e., InfoNCE loss \cite{InfoNCE}.
\emph{Our insight is to inject motif-aware hardness}, and thus the minimization of Riemannian Contrast (RC) is
\begin{equation}
\operatorname{RC}_m^0=
- \sum_{i=1}^N \log \frac{\exp(s(\boldsymbol z^{m}_i, \boldsymbol z^{0}_i) )}
{\sum_{j=1}^N \exp(h(\boldsymbol z^m_i, \boldsymbol z^0_j|\mathcal M_3) s(\boldsymbol z^{m}_i, \boldsymbol z^{0}_j) )}.
\label{contrast}
\end{equation}
Note that, Eq. (\ref{contrast}) recovers the InfoNEC with $\alpha=0$.

\begin{algorithm}[t]
        \caption{\textbf{Optimizing MotifRGC  }       } 
        \KwIn{Graph $G(\mathcal V, \mathcal E, \mathcal M, \mathbf X)$\\
            $\quad \quad \ \ \ $ \emph{MinSteps}, \emph{MaxSteps}}
        \KwOut{Riemannian graph encoder $\Phi$\\ $\quad \quad \quad \ \ $ Motif generator $G$ \\$\quad \quad \quad \ \ \ $The discriminator $D$}
Initialize $\Phi$, $G$, $D$ and Riemannian  feature $\boldsymbol x^m$;\\
\While{not converging}{   
\hfill $\rhd$ \emph{Riemannian Generative-Contrastive Learning}\\
            \For{MinSteps} {
                Generate multiple geometric views;\\
                Calculate motif-aware hardness in Eq. \ref{hard};\\
                Generate fake motifs from $G$ with Eqs. \ref{mg1}-\ref{mg2};\\
                Update $\Phi$ and $G$ according to Eq. \ref{min};\\
            }
\hfill $\rhd$ \emph{Training the Discriminator}\\
            \For{MaxSteps}{
                Sample real motifs and fake motifs generated from $G$;\\
                Update $D$ according to Eq. \ref{max};\\
            }
}
\end{algorithm}

\subsubsection{Overall Minmax Objective.} 
The overall minimization is given by incorporating the objective of Riemannian motif generator and motif-aware Riemannian contrast, 
\begin{equation}
 \min \ \mathbb E_{\mathcal S\sim G(\mathcal S;\theta_G)}[\log (1-D(\mathcal S;\theta_D))] + \sum\nolimits_{m=1}^M(\operatorname{RC}_m^0+\operatorname{RC}_0^m).
\label{min}
\end{equation}
while maximization is the objective of the discriminator,
\begin{equation}
 \max \  \mathbb E_{\mathcal M}[\log D(\mathcal M;\theta_D)] +\mathbb E_{\mathcal S\sim G(\mathcal S;\theta_G)}[\log (1-D(\mathcal S;\theta_D))].
\label{max}
\end{equation}

The overall process is summarized in Algorithm 1, 
where the minimization is given in Lines $4$-$8$ while maximization in Lines $10$-$12$. 
Line $1$ is specified in Experiment.
\emph{Alternatively optimizing the minimization in Eq. \ref{min} and maximization in Eq. \ref{max},
we learn node representations in the diverse-curvature manifold, where motif regularity is captured.}

\begin{table*}[ht]
\caption{Link prediction results on Cora, Citeseer, Pubmed and Airport datasets in terms of AUC (\%) and AP (\%). Standard derivations are given in the brackets. The best results are \textbf{boldfaced} and the runner up \underline{underlined}. }
\vspace{-0.05in}
\label{tab:lp}
\centering
\resizebox{1.02\linewidth}{!}{
\begin{tabular}{ r| cc |c c| cc|cc}
\hline
  & \multicolumn{2}{c|}{\textbf{Cora}}         & \multicolumn{2}{c|}{\textbf{Citeseer}}        & \multicolumn{2}{c|}{\textbf{Pubmed}}                     & \multicolumn{2}{c}{\textbf{Airport} }                 \\
\textbf{Method}  & \multicolumn{1}{c}{AUC} & \multicolumn{1}{c}{AP} \vline & \multicolumn{1}{c}{AUC} & \multicolumn{1}{c}{AP} \vline & \multicolumn{1}{c}{AUC} & \multicolumn{1}{c}{AP} \vline & \multicolumn{1}{c}{AUC} & \multicolumn{1}{c}{AP} \\
\hline
 GCN   
& 90.11(0.51)     & 91.52(0.68)     & 90.16(0.49)     & 92.91(0.33)   
& 91.16(0.36)     & 91.96(1.27)     & 89.29(0.38)     & 91.37(0.29) \\
 GAT 
& 92.55(0.49)     & 93.41(1.01)     & 89.32(0.36)     & 92.20(1.51)   
& 91.21(0.24)     & 92.07(1.05)     & 91.42(1.20)     & 92.26(1.06) \\
 SAGE  
& 86.02(0.55)     & 90.20(0.76)     & 88.18(0.22)     & 89.07(0.18)   
& 87.93(0.15)     & 91.42(0.12)     & 89.75(0.67)     & 92.12(0.74) \\
SME 
& 92.13(0.28)     & 93.24(0.20)     & 93.20(0.66)     & 94.56(0.92)   
& 93.11(0.09)     & 93.36(0.75)     & 92.68(0.33)     & 94.08(0.57) \\
\hline
 HGCN   
& 93.60(0.37)     & 94.13(0.27)     & 94.33(0.42)     & 94.90(0.24)   
& 95.43(0.02)     & 95.44(0.02)     & 95.11(0.26)     & 95.09(0.31) \\
HGNN  
& 91.59(1.67)     & 91.34(1.37)     & 96.05(0.42)     & 96.82(0.44)  
& 94.37(0.63)     & 93.57(1.03)     & 92.46(0.39)     & 93.36(0.82) \\
LGCN   
& 92.69(0.26)     & 93.37(0.26)     & 93.49(1.11)     & 94.32(0.77)   
& 95.40(0.20)     & 95.52(0.22)     & 95.50(0.06)     & 95.61(0.11) \\
$\kappa$-GNN   
& 94.25(1.23)     & 93.62(1.76)     & 97.06(0.67)     & 96.59(0.43)    
& 94.90(0.30)     & 94.84(0.13)     & 95.08(0.81)     & 95.21(0.80) \\
 H-to-H  
& 89.71(0.85)     & 91.70(0.45)     & 91.23(3.13)     & 92.64(2.18)  
& 97.10(0.03)     & 96.57(0.07)     & 96.91(0.03)     & 95.59(0.03) \\
SelfMG
& 94.28(0.64)     & 93.95(0.72)     & 96.52(0.36)     & 95.77(1.52)
& 97.32(0.24)     & 95.16(0.53)     & 96.85(0.53)     & 94.33(1.05)\\
 $\mathcal{Q}$-GCN   
& 95.22(0.29)     & 97.09(2.03)     & 94.31(0.73)     & 94.85(0.49)   
& 94.69(0.18)     & 95.87(0.62)     & 96.49(0.13)     & 97.01(1.12) \\
 \emph{fully}H 
& 91.81(1.92)     & 92.20(0.61)     & 91.42(0.28)     & 91.33(0.81)   
& 97.51(3.89)     & 92.55(2.71)     & 94.77(0.41)     & 94.02(0.11) \\
    \hline
\textbf{Ours}$_{\text{GCN}}$
& 97.37(2.06)     & 97.95(1.02)  & 98.54(0.67)     & 97.46(0.15)    
& \underline{98.96}(0.36) & \underline{98.84}(0.40) & 97.22(0.42)       & \underline{97.28}(0.27) \\
\textbf{Ours}$_{\text{GAT}}$
&\textbf{98.86}(1.02)   &\underline{98.64}(0.15) &\underline{98.85}(0.82) &\underline{99.02}(0.75) & \textbf{99.09}(0.13) & \textbf{98.93}(0.18)      & \underline{97.40}(0.52)     & 96.72(0.59) \\
\textbf{Ours}$_{\text{SAGE}}$ 
& \underline{98.19}(0.79) & \textbf{99.02}(0.88) & \textbf{98.99}(0.51)   & \textbf{99.12}(0.52)    
& 98.92(0.67)                     & 98.69(0.78)              & \textbf{97.90}(0.35)   & \textbf{97.62}(0.40)  \\
\hline
\end{tabular}
}
\vspace{-0.1in}
\end{table*}

\section{Experiment}


\noindent \textbf{Datasets \& Baselines $\ $}
We choose 4 public datasets: Cora, Citeseer and Pubmed \cite{yang2016revisiting}, and Airport \cite{chami2019hyperbolica}. 
%
We include $8$ supervised baselines:  
HGNN \cite{liu2019hyperbolic}, HGCN \cite{chami2019hyperbolica}, LGCN \cite{zhang2021lorentzian}, $\kappa$-GCN \cite{bachmann2020constant}, H-to-H \cite{dai2021hyperbolictohyperbolic}, \emph{fully}H \cite{chen2022fullya} and  $\mathcal{Q}$-GCN \cite{xiong2022pseudoriemannian},
  and a self-supervised SelfMG \cite{aaai22SelfMix}.
We implement $\kappa$-GCN in the product manifold, showing the result of the product without our diversified factor. 
\emph{Few study considers motif in Riemannian manifold to our best knowledge, and we bridge this gap in this paper.}
We implement \textbf{our model} with the backbone of GCN \cite{kipf2017semisupervised}, GAT \cite{velickovic2018graph} and SAGE \cite{hamilton2018inductive}, 
and thus we list the results of the $3$ backbones.
In addition, we include the recent SME \cite{ijcai22motif} as a baseline, which considers the motif in Euclidean space.


\noindent \textbf{Evaluation Metrics $\ $}
We examine our model in both link prediction and node classification. 
The evaluation metrics for link prediction is AUC and Average Precision (AP) \cite{chami2019hyperbolica}, 
while for node classification, we employ Accuracy (ACC) \cite{kipf2017semisupervised}.

\noindent\textbf{Reproducibility$\ $}
In our model, the convolution layer is stacked twice, and $\operatorname{MLP}$ has $2$ hidden layers. 
The number of learnable factors is  $3$ with the curvatures $\kappa_1=1$, $\kappa_2=-1$ and $\kappa_3=-1$ as default.
$\alpha=2$ for contrastive learning.
To initialize Riemannian features,
we first initialize $\mathbf X \in \mathbb R^{N \times d_m}$, 
where $N$ and $d_m$ are number of nodes and factor dimension.
Then, we have $\mathbf X^m= \frac{\mathbf X}{2\sqrt{\kappa_m}\|\mathbf X\|_{\operatorname{max}}} \in \mathbb G^{d_m}_{\kappa_m}$ 
in the factor manifold, where  $\|\mathbf X\|_{\operatorname{max}}$ is the maximum norm of the rows.
Riemannian features are optimized by Riemannian Adam \cite{GeoOpt}, while others are optimized by Adam \cite{iclr14adam}.



\subsection{Empirical Results}
\subsubsection{Link prediction and Node classification}

We summarize the results for link prediction and node classification in Table \ref{tab:lp} and Table \ref{tab:node_cls}, respectively.
We run each method $10$ times independently, and report the mean value with standard deviation for fair comparisons. 
For our model, the representation dimension is $128$ with $32$ for each factor, except two factors on Pubmed.
All the methods leverage Fermi-Dirac decoder \cite{nips17NickelK} for link prediction, where the distance is given by the respective geometry.
We achieve the best results among $12$ baselines, 
e.g., our model of GCN backbone achieves at least $7.36\%$ AUC gain to the backbone iteself.
The reason is that the constructed manifold better matches graph structures, and our design for motifs benefits representation and curvature learning.



\subsubsection{On Numerical Stability}
Typical Riemannian GCNs with the exponential/logarithmic map sometimes encounter ``Not a Number (NaN)'' issue in PyTorch, as will be shown in Ablation Study.
To avoid this issue,
$\kappa$-GNN carefully tunes the curvatures for the maps.
HGCN, LGCN, SelfMG and $Q$-GCN employ the post-process with Normalize function, forcing the representations to fit Riemannian manifolds.
H-to-H alternatively introduce Lorentz-type operators but the sophisticated optimization in Stiefel manifold is nontrivial.
On the contrary, the proposed $\operatorname{gF}$ is numerical stable.



\begin{table}[ht]
\caption{Node classification results in terms of ACC (\%).}
\vspace{-0.05in}
\label{tab:node_cls}
\centering
\begin{tabular}{r|c c c c}
\hline
\textbf{Method} & \multicolumn{1}{c}{\textbf{Cora}} & \multicolumn{1}{c}{\textbf{Citeseer} }& \multicolumn{1}{c}{ \textbf{Pubmed} } & \multicolumn{1}{c}{\textbf{ Airport}} \\
\hline
  GCN
 & 81.1(0.3)                     & 70.2(0.2)                  & 77.7(1.0)                     & 81.7(0.4) \\
  GAT
 & 81.9(1.1)                     & 71.3(1.4)                  & 78.6(1.3)                     & 82.1(1.2) \\
  SAGE
 & 77.9(1.7)                     & 69.3(1.9)                  & 77.2(0.5)                     & 82.1(0.9) \\
  SME
 & 81.5(0.6)                     & 72.5(1.0)                  & 78.4(0.7)                     & 83.9(0.7) \\
    \hline
  HGCN
 & 81.1(0.2)                     & 72.5(1.0)                     & 80.1(1.6)                   & 89.2(0.6) \\
  HGNN
 & 80.8(0.6)                     & 71.1(2.2)                     & 79.7(1.2)                   & 82.7(2.2) \\
  LGCN
 & \underline{82.6}(0.3) & 72.4(0.3)                     & 80.3(0.3)                   & 89.2(1.3) \\
  $\kappa$-GCN
 & 81.2(0.4)                     & 73.1(0.6)                     & 81.2(0.5)                   & 84.8(1.5) \\
  H-to-H
 & 80.0(0.2)                     & 71.0(0.3)                     & 79.5(0.3)                   & 81.8(0.9) \\
  SelfMG
 & 82.2(1.0)                     & 73.2(1.0)                     & 80.0(1.2)                   & 86.1(1.2) \\
  $Q$-GCN
 & 81.7(0.6)                     & 73.5(0.4)                     & 80.9(0.8)                   & 83.2(0.8) \\
  \emph{fully}H
 & 79.9(0.4)                     & 71.9(1.1)                     & 79.9(0.5)                   & 86.4(0.3) \\
    \hline
\textbf{Ours}$_{\text{\scriptsize{GCN}}}$
& \textbf{83.7}(1.2)       &\textbf{74.2}(1.1)       &\textbf{82.2}(1.1)       & 89.2(1.8) \\
\textbf{Ours}$_{\text{\scriptsize{GAT}}}$
& 82.5(0.7)                     &\underline{73.9}(0.7)   & 81.5(0.9)                     & \textbf{91.5}(1.3) \\
\textbf{Ours}$_{\text{\scriptsize{SAGE}}}$
& 82.0(0.5)                     & 73.3(1.1)                     &\underline{81.6}(1.0) & \underline{90.9}(1.8) \\
\hline
\end{tabular}
\vspace{-0.3in}
\end{table}

\begin{table}[htb]
\caption{Ablation Study on Proposed Components. AUC (\%) for Link Prediction (LP). ACC(\%) for Node Classification (NC). Best result are \textbf{boldfaced} and runner-up \underline{underlined}. }
    \vspace{-0.05in}
\label{lab:ablation 1}
\begin{tabular}{c l|cc|cc}
\hline
                                &                &  \multicolumn{2}{c|}{\textbf{Cora}} & \multicolumn{2}{c}{\textbf{Citeseer}}  \\
  \multicolumn{2}{c|}{\textbf{Variant}}  & LP & NC & LP & NC  \\
\hline
\multirow{5}{*}{\rotatebox{90}{GCN}} 
& $\operatorname{MotifRGC}$ 
& \textbf{97.37}             &\textbf{83.67}          &\textbf{98.54}       & \textbf{74.24} \\
& $-\operatorname{Hard}$ 
&  \underline{97.05}      &\underline{83.12}    &\underline{98.36}  & \underline{73.99}\\
& $-\operatorname{RCL}$ 
& 96.30                            & 82.34                       & 97.86                    & 71.98 \\
& $-\operatorname{gF}$ 
& 96.92                            & 82.80                       & NaN                      & 73.20 \\
& $-\operatorname{Motif}$
& 95.02                            & 81.29                       & 96.80                     & 72.16 \\
\hline
\multirow{5}{*}{\rotatebox{90}{GAT}} 
& $\operatorname{MotifRGC}$
& \textbf{98.86}             & \textbf{82.55}          & \textbf{98.85}      & \textbf{73.95}\\
& $-\operatorname{Hard}$
&  \underline{98.45}      & \underline{82.20}     & 98.27                    &\underline{73.39} \\
& $-\operatorname{RCL}$ 
& 97.99                            & 81.06                        & 97.86                     & 72.93 \\
& $-\operatorname{gF}$ 
& 98.10                            & 82.02                        & \underline{98.79} & 73.08 \\
& $-\operatorname{Motif}$
& 95.18                            & 82.13                        & 97.11                     & 72.67 \\
\hline
\multirow{5}{*}{\rotatebox{90}{SAGE}} 
& $\operatorname{MotifRGC}$
& \textbf{98.19}              & \textbf{82.02}       & \textbf{98.99}           & \textbf{73.35}\\
& $-\operatorname{Hard}$
& \underline{98.02}        &\underline{81.88}   & 98.72                         &\underline{72.86} \\
& $-\operatorname{RCL}$ 
& 96.81                            & 79.02                      & 97.15\tiny{(0.09)}     & 72.11 \\
& $-\operatorname{gF}$ 
& 97.96                            & 81.30                      & \underline{98.88}      & 72.57 \\
& $-\operatorname{Motif}$
& 94.50                            & 81.49                      & 96.67                          &72.32 \\
 \hline
\end{tabular}
    \vspace{-0.1in}
\end{table}

\begin{table}[htb]
\caption{Ablation Study on Learnable Factors.}
    \vspace{-0.05in}
\centering
\label{lab:ablation2}
\begin{tabular}{c l|cc|cc}
\hline
& &  \multicolumn{2}{c|}{\textbf{Cora}} & \multicolumn{2}{c}{\textbf{Citeseer}}  \\
  \multicolumn{2}{c|}{\textbf{Variant}}  & LP & NC & LP & NC  \\
\hline
\multirow{4}{*}{\rotatebox{90}{GCN}} 
& $\mathbb G \times \mathbb S_{\operatorname{up}}$ 
& 95.27     & 81.30       & 96.33      & 71.92  \\
& $(\mathbb G)^2 \times \mathbb S_{\operatorname{up}}$ 
& 95.80     & 82.06       & 97.46      & 73.14  \\
& $(\mathbb G)^3 \times \mathbb S_{\operatorname{up}}$ 
& \underline{97.37}  & \underline{83.67}       & \underline{98.54}     & \textbf{74.24} \\
& $(\mathbb G)^4 \times \mathbb S_{\operatorname{up}}$ 
& \textbf{97.99}     & \textbf{83.83}       & \textbf{99.03}      & \underline{74.18} \\
\hline
\multirow{4}{*}{\rotatebox{90}{GAT}} 
& $\mathbb G \times \mathbb S_{\operatorname{up}}$ 
& 95.87    & 82.09       & 97.68      & 72.60  \\
& $(\mathbb G)^2 \times \mathbb S_{\operatorname{up}}$ 
& 97.58     & 82.24       & 98.50      & 73.31  \\
& $(\mathbb G)^3 \times \mathbb S_{\operatorname{up}}$ 
& \underline{98.86}     & \textbf{82.55}       & \underline{98.85}      & \underline{73.95}\\
& $(\mathbb G)^4 \times \mathbb S_{\operatorname{up}}$ 
& \textbf{99.16}     & \underline{82.40}       & \textbf{98.96}      & \textbf{74.02}\\
\hline
\end{tabular}
    \vspace{-0.2in}
\end{table}

\subsubsection{Ablation Study}

Here, we evaluate the effectiveness of each proposed component of MotifRGC: 1) generalized Fourier map $\operatorname{gF}$, 2) Riemannian motif generator, 3) Riemannian contrastive learning, 4) motif-aware hardness as well as the number of learnable factors. 

To this end, we introduce a variety of variant models.
For $-\operatorname{gF}$ variant, we replace $\operatorname{gF}$ with logarithmic map to evaluate the numerical stability of $\operatorname{gF}$.
For $-\operatorname{motif}$ variant, we train our model with the contrastive loss only, to evaluate the importance of motif generation.
For $-\operatorname{RCL}$ variant, we disable Riemannian contrastive learning in the minmax objective.
For $-\operatorname{Hard}$ variant, we replace $\operatorname{RCL}$ with InfoNCE loss \cite{InfoNCE} in the minmax objective to evaluate the motif-aware hardness in contrastive learning.
The variants are instantiated on the product of three learnable factors by default, i.e., $(\mathbb G)^3 \times \mathbb S_{\operatorname{up}}$.
We record the results for both tasks on Cora and Citeseer datasets in Table \ref{lab:ablation 1}, where the derivation is given in brackets.
It shows that: 1) $-\operatorname{gF}$ variant has inferior performance to our model. 
The reason lies in that the logarithmic map suffers from numerical instability \cite{chen2022fullya,Yu2022RandomLF}.
The issue of ``NaN'' may occur, as shown in the link prediction on Citeseer,
 \emph{motivating our formulation of $\operatorname{gF}$.}
2) $\operatorname{RCL}$ increases the performance of our model, and motif-aware hardness is important for $\operatorname{RCL}$. 
3) Our model consistently outperforms the $-\operatorname{motif}$ variant.
\emph{It verifies our insight: motif regularity is effective to learn the curvatures},
offering a new perspective to curvature learning in Riemannian manifold.
On the number of learnable factors, we instantiate our model with different numbers of factors, and record the results in Table \ref{lab:ablation2}. 
It suggests: the product with more factor is more flexible to match the graph structures in general, achieving better results.


\subsubsection{Case Study on Triangle Generation}

\begin{figure}
\centering
    \includegraphics[width=1\linewidth]{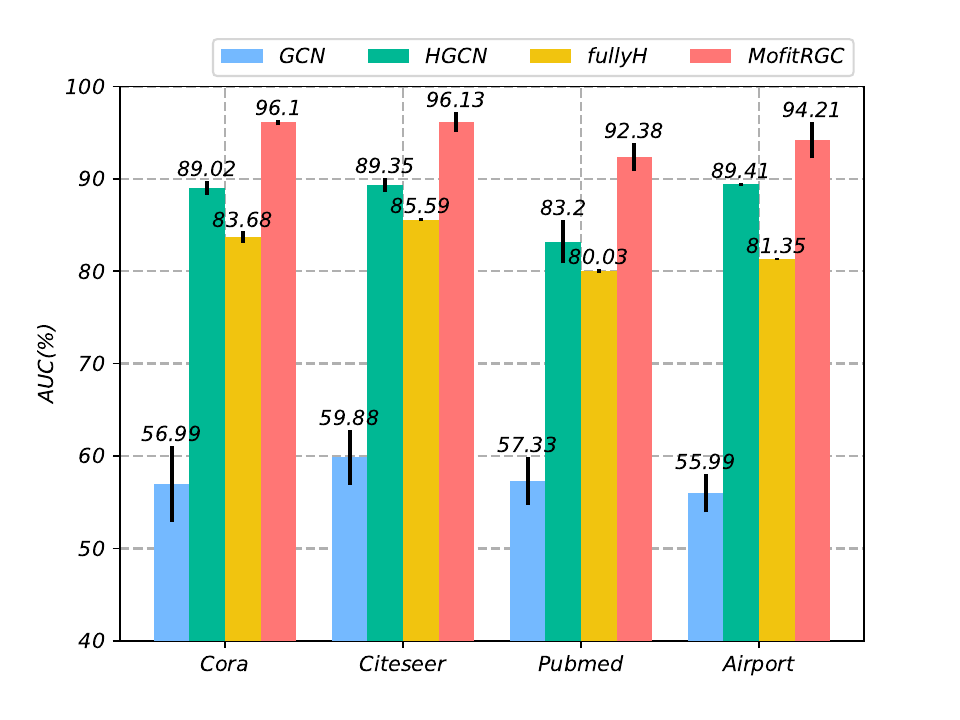}
        \vspace{-0.35in}
     \caption{AUC (\%) of Triangle Generation/Prediction.}
    \label{casestudy}
    \vspace{-0.15in}
\end{figure}

We conduct the case study to examine triangle generation from the proposed MotifRGC.
Concretely, MotifRGC is instantiated on the backbone of GCN, and the generated triangles are evaluated by AUC metric given real triangles.
Previous Riemannain methods lack the ability of motif generation, to our best knowledge.
Instead, we list the AUC of triangle prediction, where a triangle is predicted if all the three edges are correctly predicted.
Fig. \ref{casestudy} shows the result of triangle generation/prediction on Cora, Citeseer, Pubmed and Airport datasets.
Standard deviation is given in the error bar.
In Fig.  \ref{casestudy}, Riemannain models outperform the Euclidean GCN. 
It suggest that Riemannian manifold tends to be more suitable to model motifs than the Euclidean counterpart.
Our MotifRGC achieves the best result, with at least  $4.8\%$ AUC gain to the runnerup and $35.05\%$ AUC gain to its backbone.  
It shows \emph{the learnt manifold captures the motif regularity of real graphs, and thus is able to generate triangles effectively.}


\section{Related Work}

We briefly summarize the related work in Riemannian graph representation learning.
\citet{nips17NickelK,DBLP:conf/icml/0008TO19} introduce hyperbolic space to model graphs. 
HGNN \cite{liu2019hyperbolic} and HGCN \cite{chami2019hyperbolica} generalize GCN \cite{kipf2017semisupervised} and GAT \cite{velickovic2018graph}  to hyperbolic space, respectively.
\citet{zhang2021lorentzian,DBLP:conf/www/LiFSJTWP22} give other formulations of hyperbolic GNNs.
\citet{DBLP:conf/icdm/Fu0WSJWTPY21} propose to explore curvature in hyperbolic space.
H-to-H \cite{dai2021hyperbolictohyperbolic} and \emph{fully}H \cite{chen2022fullya} study the Lorentz type operations for hyperbolic GCNs.
$\kappa$-GCN \cite{bachmann2020constant} further extends GCN to  $\kappa$-sterographical model of any curvature.
$\mathcal Q$-GCN \cite{xiong2022pseudoriemannian} introduces a quotient manifold of a single curvature radius.
\citet{www22DualSpace,nips20GeoInteration} embed graphs in the dual spaces of Euclidean and hyperbolic ones.
\citet{aaai21RiemmanMatrix,icml21SymmSpace} study Riemannian matrix spaces for graphs.
\citet{DBLP:conf/kdd/00010PK23,www23curvDrop,icml18revisitCurv} leverage Ricci curvature to model graphs.
\citet{kdd22ultrahyperKG,www21mixedKG} focus on knowledge graphs specially, different from our setting. 
SelfMG \cite{aaai22SelfMix} and $\kappa$-GCN also study graphs in the product space \cite{nips19product} but, as mentioned before, \emph{we specify the product itself cannot result in diverse-curvature manifold.}
 \citet{digiovanni2022heterogeneous} study the curvature diversity of an upper hypersphere mathematically, but \emph{neither build GCN nor consider curvature learning.}
Note that, \citet{icml22RidgeletKernel,Yu2022RandomLF} formulate invariant kernels for hyperbolic space, while \emph{we formulate an  invariant kernel for any Riemannian manifold with theoretical guarantee.}
Recently, \citet{SunL23AAAI,SunL22CIKM,HVGNN} propose Riemannian GNNs for temporal graphs. \citet{SunL23ICDM,SunL23IJCAI} rethink structure learning and node clustering in the manifold, respectively.

\section{Conclusion}

We rethink Riemannian graph representation learning, and propose the first model considering motifs in diverse-curvature manifold.
We design a new type of Riemannian GCN (D-GCN), where we ensure curvature diversity by the product layer, and address numerical stability by the kernel layer.
Then, motif-aware Riemannian  generative-contrastive learning  introduces a minmax game in the constructed manifold, capturing motif-regularity in node representations.
Extensive experiments show the superiority of our model.

\section{Acknowledgments}
This work is supported by National Key R\&D Program of China through grant 2021YFB1714800, the National Natural Science Foundation of China through grant 62202164, and the Fundamental Research Funds for the Central Universities (2022MS018).
Prof. Philip S. Yu is supported in part by NSF under grant III-2106758. 
Corresponding Authors: Li Sun and Hao Peng.

\bibliography{aaai24}

\begin{thebibliography}{51}
\providecommand{\natexlab}[1]{#1}

\bibitem[{Bachmann, B{\'{e}}cigneul, and Ganea(2020)}]{bachmann2020constant}
Bachmann, G.; B{\'{e}}cigneul, G.; and Ganea, O. 2020.
\newblock Constant Curvature Graph Convolutional Networks.
\newblock In \emph{Proceedings of the 37th International Conference on Machine
  Learning (ICML)}, volume 119, 486--496. {PMLR}.

\bibitem[{B{\'{e}}cigneul and Ganea(2019)}]{GeoOpt}
B{\'{e}}cigneul, G.; and Ganea, O. 2019.
\newblock Riemannian Adaptive Optimization Methods.
\newblock In \emph{Proceedings of 7th International Conference on Learning
  Representation (ICLR)}.

\bibitem[{Chami et~al.(2019)Chami, Ying, R{\'{e}}, and
  Leskovec}]{chami2019hyperbolica}
Chami, I.; Ying, Z.; R{\'{e}}, C.; and Leskovec, J. 2019.
\newblock Hyperbolic Graph Convolutional Neural Networks.
\newblock In \emph{Advances in the 32nd Conference on Neural Information
  Processing Systems (NeurIPS)}, 4869--4880.

\bibitem[{Chen et~al.(2022)Chen, Han, Lin, Zhao, Liu, Li, Sun, and
  Zhou}]{chen2022fullya}
Chen, W.; Han, X.; Lin, Y.; Zhao, H.; Liu, Z.; Li, P.; Sun, M.; and Zhou, J.
  2022.
\newblock Fully Hyperbolic Neural Networks.
\newblock In \emph{Proceedings of the 60th Annual Meeting of the Association
  for Computational Linguistics (ACL)}, 5672--5686. ACL.

\bibitem[{Cruceru, B{\'{e}}cigneul, and Ganea(2021)}]{aaai21RiemmanMatrix}
Cruceru, C.; B{\'{e}}cigneul, G.; and Ganea, O. 2021.
\newblock Computationally Tractable Riemannian Manifolds for Graph Embeddings.
\newblock In \emph{Proceedings of the 35th AAAI Conference on Artificial
  Intelligence (AAAI)}, 7133--7141. AAAI Press.

\bibitem[{Dai et~al.(2021)Dai, Wu, Gao, and
  Jia}]{dai2021hyperbolictohyperbolic}
Dai, J.; Wu, Y.; Gao, Z.; and Jia, Y. 2021.
\newblock A Hyperbolic-to-Hyperbolic Graph Convolutional Network.
\newblock In \emph{Proceedings of the IEEE/CVF Conference on Computer Vision
  and Pattern Recognition (CVPR)}, 154--163. CVF/ {IEEE}.

\bibitem[{Duval and Malliaros(2022)}]{HostPool}
Duval, A.; and Malliaros, F.~D. 2022.
\newblock Higher-order Clustering and Pooling for Graph Neural Networks.
\newblock In \emph{Proceedings of the 31st ACM International Conference on
  Information and Knowledge Management (CIKM)}, 426--435.

\bibitem[{Fu et~al.(2021)Fu, Li, Wu, Sun, Ji, Wang, Tan, Peng, and
  Yu}]{DBLP:conf/icdm/Fu0WSJWTPY21}
Fu, X.; Li, J.; Wu, J.; Sun, Q.; Ji, C.; Wang, S.; Tan, J.; Peng, H.; and Yu,
  P.~S. 2021.
\newblock {ACE-HGNN:} Adaptive Curvature Exploration Hyperbolic Graph Neural
  Network.
\newblock In \emph{Proceedings of the 21st {IEEE} International Conference on
  Data Mining (ICDM)}, 111--120. {IEEE}.

\bibitem[{Giovanni, Luise, and Bronstein(2022)}]{digiovanni2022heterogeneous}
Giovanni, F.~D.; Luise, G.; and Bronstein, M.~M. 2022.
\newblock Heterogeneous manifolds for curvature-aware graph embedding.
\newblock In \emph{Proceedings of the 10th International Conference on Learning
  Representation (GTRL Workshop)}.

\bibitem[{Gu et~al.(2019)Gu, Sala, Gunel, and R{\'{e}}}]{nips19product}
Gu, A.; Sala, F.; Gunel, B.; and R{\'{e}}, C. 2019.
\newblock Learning Mixed-Curvature Representations in Product Spaces.
\newblock In \emph{Proceedings of the 7th International Conference on Learning
  Representation (ICLR)}.

\bibitem[{Hamilton, Ying, and Leskovec(2017)}]{hamilton2018inductive}
Hamilton, W.~L.; Ying, Z.; and Leskovec, J. 2017.
\newblock Inductive Representation Learning on Large Graphs.
\newblock In \emph{Advances in the 30th Conference on Neural Information
  Processing Systems (NeurIPS)}, 1024--1034.

\bibitem[{Jiang et~al.(2022)Jiang, Yang, Wen, Su, and Huang}]{ijcai22motif}
Jiang, X.; Yang, Z.; Wen, P.; Su, L.; and Huang, Q. 2022.
\newblock A Sparse-Motif Ensemble Graph Convolutional Network against
  Over-smoothing.
\newblock In \emph{Proceedings of the 31st International Joint Conference on
  Artificial Intelligence (IJCAI)}, 2094--2100.

\bibitem[{Kingma and Ba(2015)}]{iclr14adam}
Kingma, D.~P.; and Ba, J. 2015.
\newblock Adam: {A} Method for Stochastic Optimization.
\newblock In \emph{Proceedings of the 3rd International Conference on Learning
  Representation (ICLR)}.

\bibitem[{Kipf and Welling(2017)}]{kipf2017semisupervised}
Kipf, T.~N.; and Welling, M. 2017.
\newblock Semi-Supervised Classification with Graph Convolutional Networks.
\newblock In \emph{Proceedings of the 5th International Conference on Learning
  Representation (ICLR)}.

\bibitem[{Krioukov et~al.(2010)Krioukov, Papadopoulos, Kitsak, Vahdat, and
  Bogu{\~{n}}{\'{a}}}]{DBLP:journals/corr/abs-1006-5169}
Krioukov, D.~V.; Papadopoulos, F.; Kitsak, M.; Vahdat, A.; and
  Bogu{\~{n}}{\'{a}}, M. 2010.
\newblock Hyperbolic Geometry of Complex Networks.
\newblock \emph{CoRR}, abs/1006.5169.

\bibitem[{Li et~al.(2022{\natexlab{a}})Li, Fu, Sun, Ji, Tan, Wu, and
  Peng}]{www22curvGAN}
Li, J.; Fu, X.; Sun, Q.; Ji, C.; Tan, J.; Wu, J.; and Peng, H.
  2022{\natexlab{a}}.
\newblock Curvature Graph Generative Adversarial Networks.
\newblock In \emph{Proceedings of The {ACM} Web Conference 2022}, 1528--1537.
  {ACM}.

\bibitem[{Li et~al.(2022{\natexlab{b}})Li, Fu, Sun, Ji, Tan, Wu, and
  Peng}]{DBLP:conf/www/LiFSJTWP22}
Li, J.; Fu, X.; Sun, Q.; Ji, C.; Tan, J.; Wu, J.; and Peng, H.
  2022{\natexlab{b}}.
\newblock Curvature Graph Generative Adversarial Networks.
\newblock In \emph{Proceedings of the {ACM} Web Conference 2022}, 1528--1537.
  {ACM}.

\bibitem[{Liu and Sariy{\"{u}}ce(2023)}]{DBLP:conf/kdd/LiuS23}
Liu, P.; and Sariy{\"{u}}ce, A.~E. 2023.
\newblock Using Motif Transitions for Temporal Graph Generation.
\newblock In \emph{Proceedings of the 29th SIGKDD Conference on Knowledge
  Discovery and Data Mining (KDD)}, 1501--1511. {ACM}.

\bibitem[{Liu, Nickel, and Kiela(2019)}]{liu2019hyperbolic}
Liu, Q.; Nickel, M.; and Kiela, D. 2019.
\newblock Hyperbolic Graph Neural Networks.
\newblock In \emph{Advances in the 32nd Conference on Neural Information
  Processing Systems (NeurIPS)}, 8228--8239.

\bibitem[{Liu et~al.(2023)Liu, Zhou, Pan, Wu, Li, Chen, and
  Zhang}]{www23curvDrop}
Liu, Y.; Zhou, C.; Pan, S.; Wu, J.; Li, Z.; Chen, H.; and Zhang, P. 2023.
\newblock CurvDrop: {A} Ricci Curvature Based Approach to Prevent Graph Neural
  Networks from Over-Smoothing and Over-Squashing.
\newblock In \emph{Proceedings of the {ACM} Web Conference 2023}, 221--230.
  {ACM}.

\bibitem[{L{\'{o}}pez et~al.(2021)L{\'{o}}pez, Pozzetti, Trettel, Strube, and
  Wienhard}]{icml21SymmSpace}
L{\'{o}}pez, F.; Pozzetti, B.; Trettel, S.; Strube, M.; and Wienhard, A. 2021.
\newblock Symmetric Spaces for Graph Embeddings: {A} Finsler-Riemannian
  Approach.
\newblock In \emph{Proceedings of the 38th International Conference on Machine
  Learning (ICML)}, volume 139, 7090--7101. {PMLR}.

\bibitem[{Nguyen et~al.(2023)Nguyen, Nong, Nguyen, Ho, Osher, and
  Nguyen}]{icml18revisitCurv}
Nguyen, K.; Nong, H.; Nguyen, V.; Ho, N.; Osher, S.; and Nguyen, T. 2023.
\newblock Revisiting Over-smoothing and Over-squashing Using Ollivier-Ricci
  Curvature.
\newblock In \emph{Proceedings of the 40th International Conference on Machine
  Learning (ICML)}. {PMLR}.

\bibitem[{Nickel and Kiela(2017)}]{nips17NickelK}
Nickel, M.; and Kiela, D. 2017.
\newblock Poincar{\'{e}} Embeddings for Learning Hierarchical Representations.
\newblock In \emph{Advances in 30th Conference on Neural Information Processing
  Systems (NeurIPS)}, 6338--6347.

\bibitem[{Oord, Li, and Vinyals(2018)}]{InfoNCE}
Oord, A. v.~d.; Li, Y.; and Vinyals, O. 2018.
\newblock Representation Learning with Contrastive Predictive Coding.
\newblock \emph{CoRR}, abs/1807.03748.

\bibitem[{Perozzi, Al{-}Rfou, and Skiena(2014)}]{kdd14DeepWalk}
Perozzi, B.; Al{-}Rfou, R.; and Skiena, S. 2014.
\newblock DeepWalk: Online learning of social representations.
\newblock In \emph{Proceedings of the 20th SIGKDD Conference on Knowledge
  Discovery and Data Mining (KDD)}, 701--710. {ACM}.

\bibitem[{Petersen(2016)}]{petersen2016riemannian}
Petersen, P. 2016.
\newblock \emph{Riemannian Geometry}, volume 171 of \emph{Graduate Texts in
  Mathematics}.
\newblock {Springer International Publishing}.

\bibitem[{Rahimi and Recht(2007)}]{RahimiR07-Thm}
Rahimi, A.; and Recht, B. 2007.
\newblock Random Features for Large-Scale Kernel Machines.
\newblock In \emph{Advances in 20th Conference on Neural Information Processing
  Systems (NeurIPS)}, 1177--1184.

\bibitem[{Shimizu, Mukuta, and Harada(2021)}]{iclr21hnn++}
Shimizu, R.; Mukuta, Y.; and Harada, T. 2021.
\newblock Hyperbolic Neural Networks++.
\newblock In \emph{Proceedings of 9th International Conference on Learning
  Representation (ICLR)}.

\bibitem[{Sonoda, Ishikawa, and Ikeda(2022)}]{icml22RidgeletKernel}
Sonoda, S.; Ishikawa, I.; and Ikeda, M. 2022.
\newblock Fully-Connected Network on Noncompact Symmetric Space and Ridgelet
  Transform based on Helgason-Fourier Analysis.
\newblock In \emph{Proceedings of the 39th International Conference on Machine
  Learning (ICML)}, volume 162, 20405--20422. {PMLR}.

\bibitem[{Subramonian(2021)}]{DBLP:conf/aaai/Subramonian21}
Subramonian, A. 2021.
\newblock Motif-Driven Contrastive Learning of Graph Representations.
\newblock In \emph{Proceedings of the 35th {AAAI} Conference on Artificial
  Intelligence (AAAI)}, 15980--15981. {AAAI} Press.

\bibitem[{Sun et~al.(2023{\natexlab{a}})Sun, Huang, Wu, Ye, Peng, Yu, and
  Yu}]{SunL23ICDM}
Sun, L.; Huang, Z.; Wu, H.; Ye, J.; Peng, H.; Yu, Z.; and Yu, P.~S.
  2023{\natexlab{a}}.
\newblock DeepRicci: Self-supervised Graph Structure-Feature Co-Refinement for
  Alleviating Over-squashing.
\newblock In \emph{Proceedings of the 23rd IEEE International Conference on
  Data Mining (ICDM)}.

\bibitem[{Sun et~al.(2023{\natexlab{b}})Sun, Wang, Ye, Peng, and
  Yu}]{SunL23IJCAI}
Sun, L.; Wang, F.; Ye, J.; Peng, H.; and Yu, P.~S. 2023{\natexlab{b}}.
\newblock Congregate: Contrastive Graph Clustering in Curvature Spaces.
\newblock In \emph{Proceedings of the 32nd International Joint Conference on
  Artificial Intelligence (IJCAI)}, 2296--2305.

\bibitem[{Sun et~al.(2023{\natexlab{c}})Sun, Ye, Peng, Wang, and
  Yu}]{SunL23AAAI}
Sun, L.; Ye, J.; Peng, H.; Wang, F.; and Yu, P.~S. 2023{\natexlab{c}}.
\newblock Self-Supervised Continual Graph Learning in Adaptive Riemannian
  Spaces.
\newblock In \emph{Proceedings of the 37th {AAAI} Conference on Artificial
  Intelligence (AAAI)}, 4633--4642.

\bibitem[{Sun et~al.(2022{\natexlab{a}})Sun, Ye, Peng, and Yu}]{SunL22CIKM}
Sun, L.; Ye, J.; Peng, H.; and Yu, P.~S. 2022{\natexlab{a}}.
\newblock A Self-supervised Riemannian {GNN} with Time Varying Curvature for
  Temporal Graph Learning.
\newblock In \emph{Proceedings of the 31st {ACM} International Conference on
  Information and Knowledge Management (CIKM)}, 1827--1836. {ACM}.

\bibitem[{Sun et~al.(2022{\natexlab{b}})Sun, Zhang, Ye, Peng, Zhang, Su, and
  Yu}]{aaai22SelfMix}
Sun, L.; Zhang, Z.; Ye, J.; Peng, H.; Zhang, J.; Su, S.; and Yu, P.~S.
  2022{\natexlab{b}}.
\newblock A Self-Supervised Mixed-Curvature Graph Neural Network.
\newblock In \emph{Proceedings of the 36th AAAI Conference on Artificial
  Intelligence (AAAI)}, 4146--4155. AAAI Press.

\bibitem[{Sun et~al.(2021)Sun, Zhang, Zhang, Wang, Peng, Su, and Yu}]{HVGNN}
Sun, L.; Zhang, Z.; Zhang, J.; Wang, F.; Peng, H.; Su, S.; and Yu, P.~S. 2021.
\newblock Hyperbolic Variational Graph Neural Network for Modeling Dynamic
  Graphs.
\newblock In \emph{Proceedings of the 35th {AAAI} Conference on Artificial
  Intelligence (AAAI)}, 4375--4383.

\bibitem[{Suzuki, Takahama, and Onoda(2019)}]{DBLP:conf/icml/0008TO19}
Suzuki, R.; Takahama, R.; and Onoda, S. 2019.
\newblock Hyperbolic Disk Embeddings for Directed Acyclic Graphs.
\newblock In \emph{Proceedings of the 36th International Conference on Machine
  Learning (ICML)}, volume~97, 6066--6075. {PMLR}.

\bibitem[{Topping et~al.(2022)Topping, Giovanni, Chamberlain, Dong, and
  Bronstein}]{iclr22understanding}
Topping, J.; Giovanni, F.~D.; Chamberlain, B.~P.; Dong, X.; and Bronstein,
  M.~M. 2022.
\newblock Understanding over-squashing and bottlenecks on graphs via curvature.
\newblock In \emph{Proceedings of the 10th International Conference on Learning
  Representation (ICLR)}.

\bibitem[{Velickovic et~al.(2018)Velickovic, Cucurull, Casanova, Romero,
  Li{\`{o}}, and Bengio}]{velickovic2018graph}
Velickovic, P.; Cucurull, G.; Casanova, A.; Romero, A.; Li{\`{o}}, P.; and
  Bengio, Y. 2018.
\newblock Graph Attention Networks.
\newblock In \emph{Proceedings of the 6th International Conference on Learning
  Representation (ICLR)}.

\bibitem[{Wang et~al.(2018)Wang, Wang, Wang, Zhao, Zhang, Zhang, Xie, and
  Guo}]{aaai18GraphGAN}
Wang, H.; Wang, J.; Wang, J.; Zhao, M.; Zhang, W.; Zhang, F.; Xie, X.; and Guo,
  M. 2018.
\newblock GraphGAN: Graph Representation Learning With Generative Adversarial
  Nets.
\newblock In \emph{Proceedings of the 32nd AAAI Conference on Artificial
  Intelligence (AAAI)}, 2508--2515. AAAI Press.

\bibitem[{Wang et~al.(2021)Wang, Wei, dos Santos, Wang, Nallapati, Arnold,
  Xiang, Yu, and Cruz}]{www21mixedKG}
Wang, S.; Wei, X.; dos Santos, C.~N.; Wang, Z.; Nallapati, R.; Arnold, A.~O.;
  Xiang, B.; Yu, P.~S.; and Cruz, I.~F. 2021.
\newblock Mixed-Curvature Multi-Relational Graph Neural Network for Knowledge
  Graph Completion.
\newblock In \emph{Proceedings of the Web Conference 2021}, 1761--1771. {ACM} /
  {IW3C2}.

\bibitem[{Xiong et~al.(2022{\natexlab{a}})Xiong, Zhu, Nayyeri, Xu, Pan, Zhou,
  and Staab}]{kdd22ultrahyperKG}
Xiong, B.; Zhu, S.; Nayyeri, M.; Xu, C.; Pan, S.; Zhou, C.; and Staab, S.
  2022{\natexlab{a}}.
\newblock Ultrahyperbolic Knowledge Graph Embeddings.
\newblock In \emph{Proceedings of the 28th SIGKDD Conference on Knowledge
  Discovery and Data Mining (KDD)}, 2130--2139. {ACM}.

\bibitem[{Xiong et~al.(2022{\natexlab{b}})Xiong, Zhu, Potyka, Pan, Zhou, and
  Staab}]{xiong2022pseudoriemannian}
Xiong, B.; Zhu, S.; Potyka, N.; Pan, S.; Zhou, C.; and Staab, S.
  2022{\natexlab{b}}.
\newblock Pseudo-Riemannian Graph Convolutional Networks.
\newblock In \emph{Advances in the 35th Conference on Neural Information
  Processing Systems (NeurIPS)}.

\bibitem[{Yang et~al.(2022)Yang, Chen, Pan, Li, Yu, and Xu}]{www22DualSpace}
Yang, H.; Chen, H.; Pan, S.; Li, L.; Yu, P.~S.; and Xu, G. 2022.
\newblock Dual Space Graph Contrastive Learning.
\newblock In \emph{Proceedings of the {ACM} Web Conference 2022}, 1238--1247.
  {ACM}.

\bibitem[{Yang et~al.(2023{\natexlab{a}})Yang, Zhou, Pan, and
  King}]{DBLP:conf/kdd/00010PK23}
Yang, M.; Zhou, M.; Pan, L.; and King, I. 2023{\natexlab{a}}.
\newblock {\(\kappa\)}HGCN: Tree-likeness Modeling via Continuous and Discrete
  Curvature Learning.
\newblock In \emph{Proceedings of the 29th SIGKDD Conference on Knowledge
  Discovery and Data Mining (KDD)}, 2965--2977. {ACM}.

\bibitem[{Yang et~al.(2023{\natexlab{b}})Yang, Zhou, Ying, Chen, and
  King}]{icml18revisitHyperbolic}
Yang, M.; Zhou, M.; Ying, R.; Chen, Y.; and King, I. 2023{\natexlab{b}}.
\newblock Hyperbolic Representation Learning: Revisiting and Advancing.
\newblock In \emph{Proceedings of the 40th International Conference on Machine
  Learning (ICML)}. {PMLR}.

\bibitem[{Yang, Cohen, and Salakhutdinov(2016)}]{yang2016revisiting}
Yang, Z.; Cohen, W.~W.; and Salakhutdinov, R. 2016.
\newblock Revisiting Semi-Supervised Learning with Graph Embeddings.
\newblock In Balcan, M.; and Weinberger, K.~Q., eds., \emph{Proceedings of the
  33rd International Conference on Machine Learning (ICML)}, volume~48, 40--48.
  JMLR.org.

\bibitem[{Yu and Sa(2023)}]{Yu2022RandomLF}
Yu, T.; and Sa, C.~D. 2023.
\newblock Random Laplacian Features for Learning with Hyperbolic Space.
\newblock In \emph{Proceedings of the 11th International Conference on Learning
  Representation (ICLR)}, 1--23.

\bibitem[{Yu and Gao(2022)}]{DBLP:conf/icml/YuG22}
Yu, Z.; and Gao, H. 2022.
\newblock Molecular Representation Learning via Heterogeneous Motif Graph
  Neural Networks.
\newblock In \emph{Proceedings of the 39th International Conference on Machine
  Learning (ICML)}, volume 162, 25581--25594. {PMLR}.

\bibitem[{Zhang et~al.(2021)Zhang, Wang, Shi, Liu, and
  Song}]{zhang2021lorentzian}
Zhang, Y.; Wang, X.; Shi, C.; Liu, N.; and Song, G. 2021.
\newblock Lorentzian Graph Convolutional Networks.
\newblock In \emph{Proceedings of the Web Conference 2021}, 1249--1261. {ACM} /
  {IW3C2}.

\bibitem[{Zhu et~al.(2020)Zhu, Pan, Zhou, Wu, Cao, and
  Wang}]{nips20GeoInteration}
Zhu, S.; Pan, S.; Zhou, C.; Wu, J.; Cao, Y.; and Wang, B. 2020.
\newblock Graph Geometry Interaction Learning.
\newblock In \emph{Advances in the 33rd Conference on Neural Information
  Processing Systems (NeurIPS)}.

\end{thebibliography}

\end{document}